# Age and Gender Prediction using Deep CNNs and Transfer Learning

Vikas Sheoran[1], Shreyansh Joshi[2] and Tanisha R. Bhayani[3]

[1] Birla Institute of Technology & Science, Pilani - Hyderabad campus,
Hyderabad 500078, India.
`f20180847@hyderabad.bits-pilani.ac.in`

[2] Birla Institute of Technology & Science, Pilani - Goa campus,
Goa 403726, India.
`f20180097@goa.bits-pilani.ac.in`

[3] Silver Touch Technologies Limited, Ahmedabad, 380006, India.
`t.bhayani@yahoo.com`.

**Abstract.** The last decade or two has witnessed a boom of images. With the increasing ubiquity of cameras and with the advent of selfies, the number of facial images available in the world has skyrocketed. Consequently, there has been a growing interest in automatic age and gender prediction of a person using facial images. We in this paper focus on this challenging problem. Specifically, this paper focuses on age estimation, age classification and gender classification from still facial images of an individual. We train different models for each problem and we also draw comparisons between building a custom CNN (Convolutional Neural Network) architecture and using various CNN architectures as feature extractors, namely VGG16 pre-trained on VGGFace, ResNet50 and SE-ResNet50 pre-trained on VGGFace2 dataset and training over those extracted features. We also provide baseline performance of various machine learning algorithms on the feature extraction which gave us the best results. It was observed that even simple linear regression trained on such extracted features outperformed training CNN, ResNet50 and ResNeXt50 from scratch for age estimation.

## 1 Introduction

Age and gender prediction has become one of the more recognized fields in deep learning, due to the increased rate of image uploads on the internet in today's data driven world. Humans are inherently good at determining one's gender, recognizing each other and making judgements about ethnicity but age estimation still remains a formidable problem. To emphasize more on the difficulty of the problem, consider this - the most common metric used for evaluating age prediction of a person is mean absolute error (MAE). A study reported that humans can predict the age of a person above 15 years of age with a MAE of 7.2-7.4 depending on the database conditions [1]. This means that on average, humans make predictions off by 7.2-7.4 years. The question is, can we do better? Can we automate this problem in a bid to reduce human dependency and to simultaneously obtain better results?

One must acknowledge that aging of face is not only determined by genetic factors but it is also influenced by lifestyle, expression, and environment [1]. Different people of similar age can look very different due to these reasons. That is why predicting age is such a challenging task inherently. The non-linear relationship between facial images and age/gender coupled with the huge paucity of large and balanced datasets with correct labels further contribute to this problem. Very few such datasets exist, majority datasets available for the task are highly imbalanced with a huge chunk of people lying in the age group of 20 to 75 [3]-[5] or are biased towards one of the genders. Use of such biased datasets is not prudent as it would create a distribution mismatch when deployed for testing on real-time images, thereby giving poor results.

This field of study has a huge amount of underlying potential. There has been an ever-growing interest in automatic age and gender prediction because of the huge potential it has in various fields of computer science such as HCI (Human Computer Interaction). Some of the potential applications include forensics, law enforcement [1], and security control [1]. Another very practical application involves incorporating these models into IoT. For example, a restaurant can change its theme by estimating the average age or gender of people that have entered so far.

The remaining part of the paper is organized as follows. Section 2 talks about the background and work done before in this field and how it inspired us to work. Section 3 contains the exact technical details of the project and is further divided into three subsections. Section 4 talks about the evaluation metric used. Section 5 presents the various experiments we performed along with the results we obtained, and finally section 6 wraps up the paper with conclusion and future work.

## 2    Related Work

Initial works of age and gender prediction involved techniques based on ratios of different measurements of facial features such as size of eye, nose, distance of chin from forehead, distance between the ears, angle of inclination, angle between locations [8]. Such methods were known as anthropometric methods.

Early methods were based on manual extraction of features such as PCA, LBP, Gabor, LDA, SFP. These extracted features were then fed to classical ML models such as SVMs, decision trees, logistic regression. Hu et al. [9] used the method of ULBP, PCA & SVM for age estimation. Guo et al. [10] proposed a locally adjusted robust regression (LARR) algorithm, which combines SVM and SVR when estimating age by first using SVR to estimate a global age range, and then using SVM to perform exact age estimation. The obvious down-side of such methods was that not only getting anthropometric measurements was difficult but the models were not able to generalize because people of different age and gender could have the same anthropometric measurements.

Recently the use of CNN for age and gender prediction has been widely adopted as CNNs are pretty robust and give outstanding results when tested on face images with occlusion, tilt, altered brightness. Such results have been attributed to its good ability to extract features. This happens by convolving over the given image to generate invariant features which are passed onto the next layer in a sequential fashion. It is this continual passing of information from one layer to the next that leads to CNNs being so robust and supple to occlusions, brightness changes etc.

The first application of CNNs was the Le-Net-5 [11]. However, the actual boom in using CNNs for age and gender prediction started after D-CNN [12] was introduced for image classification tasks. Rothe et al. [13] proposed DEX: Deep EXpectation of Apparent Age for age classification using an ensemble of 20 networks on the cropped faces of IMDB-Wiki dataset. Another popular work includes combining features from Deep CNN to features obtained from PCA done by Wang et al. [14].

## 3 Methodology

### 3.1 Dataset

In this paper, we use the UTKFace dataset [2] (aligned and cropped) consists of over 20,000 face images with annotations of age, gender, and ethnicity. It has a total of 23708 images of which 6 were missing age labels. The images cover large variations in facial expression, illumination, pose, resolution and occlusion. We chose this dataset because of its relatively more uniform distributions, the diversity it has in image characteristics such as brightness, occlusion and position and also because it involves images of the general public.

Some sample images from the UTKFace dataset can be seen in Fig. 1. Each image is labeled with a 3-element tuple, with age (in years), gender (Male-0, Female-1) and races (White-0, Black-1, Asian-2, Indian-3 and Others-4) respectively.

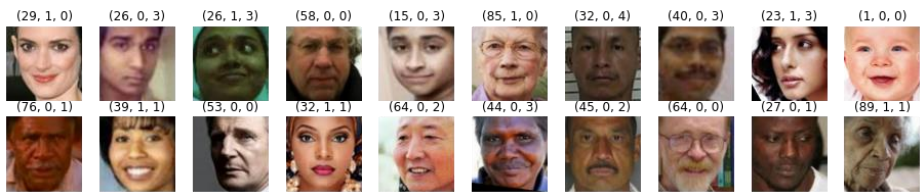

**Fig. 1.** Sample images from the UTKFace dataset.

For both our approaches (custom CNN and transfer learning based models), we used the same set of images for training, testing and validation, to have standardized results.

This was done by dividing the data sets into train, test and validation in 80 : 10 : 10 ratios. This division was done while ensuring that the data distribution in each division remains roughly the same, so that there is no distribution mismatch while training and testing the models. The Table 1 and Table 2 show the composition of training, validation and test data with respect to gender and age respectively.

4**Table 1.** Composition of sets by gender

| Gender | Training | Validation | Test | Total |
|---|---|---|---|---|
| Male | 9900 | 1255 | 1234 | 12389 |
| Female | 9061 | 1115 | 1137 | 11313 |
| Total | 18961 | 2370 | 2371 | 23702 |

**Table 2.** Composition of sets by age

| Age Group | Training | Validation | Test | Total |
|---|---|---|---|---|
| 0-10 | 2481 | 303 | 278 | 3062 |
| 11-20 | 1222 | 150 | 158 | 1530 |
| 21-30 | 5826 | 765 | 753 | 7344 |
| 31-40 | 3618 | 462 | 456 | 4536 |
| 41-50 | 1767 | 223 | 254 | 2244 |
| 51-60 | 1858 | 214 | 226 | 2298 |
| 61-70 | 1057 | 137 | 122 | 1316 |
| 71-80 | 577 | 57 | 65 | 699 |
| 81-90 | 413 | 45 | 46 | 504 |
| 91-100 | 114 | 11 | 12 | 137 |
| 101-116 | 28 | 3 | 1 | 32 |
| Total | 18961 | 2370 | 2371 | 23702 |

### 3.2 Deep CNNs

**Network Architecture.** The tasks tackled using the deep CNN approach include age and gender classification and age estimation. The basic structure of each of the 3 models includes a series of convolutional blocks, followed by a set of FC (fully connected) layers for classification and regression. An RGB image is fed to the model and is resized to 180 x 180 x 3. Every architecture comprises convolutional blocks that are a stack of convolutional layers (filter size is 3x3) followed by non-linear activation 'ReLU', max pooling (2x2) and batch normalization to mitigate the problem of covariate shift. The deeper layers here also have spatial dropout (drop value of 0.15-0.2) which drops entire feature maps to promote independence between them. Following the convolutional blocks, the output is flattened before feeding that into FC layers. These FC layers have activation function of ReLU, dropout (value between 0.2 & 0.4) and batch normalization. Table 3 shows the architecture used for age estimation.

The architectures for age classification and gender classification differ in the fact that they have 3 & 2 blocks with 256 filters respectively (in convolutional layer) and the output layer has 5 and 2 neurons respectively with softmax activation function (being classification tasks).

**Table 3.** Network Architecture for Age Estimation

| Layer | Filters | Output Size | Kernel Size | Activation |
|---|---|---|---|---|
| Image | - | 180 x 180 x 3 | - | - |
| Separable Conv1 | 64 | 180 x 180 x 64 | 3 x 3 | ReLU |
| Max Pooling | - | 90 x 90 x 64 | 2 x 2 | - |
| Separable Conv2 | 128 | 90 x 90 x 128 | 3 x 3 | ReLU |
| Max Pooling | - | 45 x 45 x 128 | 2 x 2 | - |
| Separable Conv3 | 128 | 45 x 45 x 128 | 3 x 3 | ReLU |
| Max Pooling | - | 22 x 22 x 128 | 2 x 2 | - |
| Separable Conv4 | 256 | 22 x 22 x 256 | 3 x 3 | ReLU |
| Max Pooling | - | 11 x 11 x 256 | 2 x 2 | - |
| Separable Conv5 | 256 | 11 x 11 x 256 | 3 x 3 | ReLU |
| Max Pooling | - | 5 x 5 x 256 | 2 x 2 | - |
| FC1 | - | 128 | - | ReLU |
| FC2 | - | 64 | - | ReLU |
| FC3 | - | 32 | - | ReLU |
| Output | - | 1 | - | ReLU |

**Training & Testing.** For age classification, our model classifies ages into 5 groups (0-24, 25-49, 50-74, 75-99, and 100-124). For this, we had to perform integer division (by 25) on the age values and later one hot encodes them before feeding them into the model. Similarly, gender also had to be one hot encoded for gender classification into male and female. The loss function chosen for age estimation was mean-squared error (MSE) as it is a regression task, whereas for age and gender classification it was categorical-cross entropy. For training, each model was trained using a custom data generator that allows training in mini-batches. Learning rate decay was used during training as it allowed the learning rate to decrease after a fixed number of epochs. This is essential as the learning rate becomes very precarious during the latter stages of training, when approaching convergence. Various experiments with different optimizers were conducted, the results of which have been summarized in section 5.

Each model was trained between 30 to 50 epochs on average. Initial learning rate was set of the order 1e-3. Batch size of 32 was used. The learning rate was changed to 0.6 times the current learning rate after about 9 epochs (on average) to ensure that by the end of training, the learning rate is small enough for the model to converge to the local minimum. Fig. 4 showcases the training plots of our models. In all graphs, the blue line denotes the training and the red line denotes the validation result. It is very evident that the training for gender classification was the noisiest whereas the training for age estimation was the smoothest.

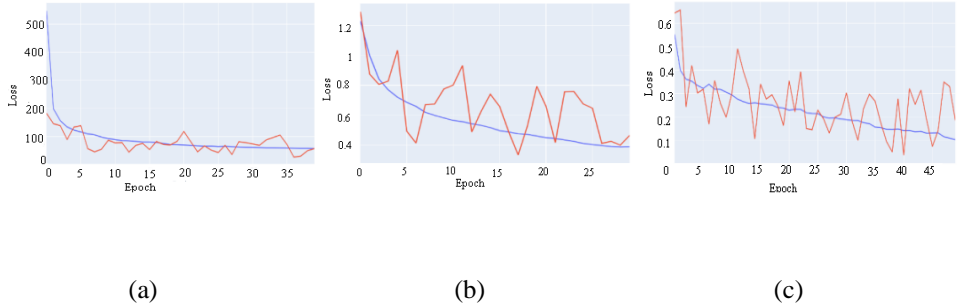

| (a) | (b) | (c) |

**Fig. 2.** Training plots depicting loss by the epoch of  a) Age Estimation  b) Age Classification  c) Gender Classification

Table 4 shows the lowest loss value to which our model could converge while training.

**Table 4.** Minimum loss value

| Age Estimation | | Age Classification | | Gender Classification | |
| --- | --- | --- | --- | --- | --- |
| Train | Validation | Train | Validation | Train | Validation |
| 56.8346 | 56.7397 | 0.3886 | 0.4009 | 0.1037 | 0.1864 |

The next subsection explores our work using transfer learning.

### 3.3   Transfer Learning

Transfer learning is one of the most powerful ideas in deep learning. It allows knowledge learned on one task to be applied to another. A lot of low-level features that determine the basic structure of the object can be very well learned from a bigger available dataset and knowledge of those transferred low-level features can help learn faster and improve performance on limited data by reducing generalization error.

The UTKFace dataset is a very small dataset to capture the complexity involved in age and gender estimation, so we focused our attention further on leveraging transfer learning. One study [6] has already compared performance of fine-tuning and pre-training state-of-the-art models for ILSVRC for age estimation on UTKFace. We take it a step further by using convolutional blocks of VGG16 pretrained on VGGFace [4] and ResNet50 and SE-ResNet-50 (SENet50 in short) pre-trained on VGGFace2 [5], as feature extractors. These models are originally proposed for facial recognition, thus can be used for higher level of feature extraction. To avoid any confusion, in this paper we denote these models as VGG_f, ResNet50_f and SENet50_f respectively where f denotes pre-trained using facial images of respective datasets.

**Network Architecture.** The tasks tackled using this transfer learning approach include age estimation and gender classification. Following is the network architecture we used in our models to train on top of features extracted.

For the gender classification, for convenience, we chose custom model names VGG_f_gender, ResNet50_f_gender and SENet50_f_gender whose design as follows. VGG_f_gender comprises of 2 blocks, each containing layers in order of batch normalization, spatial dropout with drop probability of 0.5, separable convolutions layers with 512 filters of size 3x3 with keeping padding same to reduce loss of information during convolution operations followed by max pooling with kernel size 2x2. The fully connected system consisted of batch norm layers, followed by alpha dropout, and 128 neurons with ReLU activation and He uniform initialized followed by another batch norm layer and finally the output layer with 1 neuron with sigmoid activation. Batch size chosen was 64. ResNet50_f_gender comprises of just the fully connected system with batch norm, dropout with probability of 0.5, and followed by 128 units with exponential linear units (ELU) activation, with He uniform initialization and having max norm weight constraint of magnitude 3. The output layer had single neuron with sigmoid activation. The batch size we chose for this was 128. For, SENet50_f_gender we kept the same model as for ResNet50_f_gender.

For the age estimation the models have been named VGG_f_age, ResNet50_f_age and SENet50_f_age. VGG_f_age consists of 2 convolution blocks each containing in order, a batch norm layer, spatial dropout with keep probability of 0.8 and 0.6 respectively, separable convolution layer with 512 filters of size 3x3, padding same so that dimension doesn't change (and information loss is curtailed), with ReLU activation function and He initialization. Each convolution block was followed by max pooling with kernel size 2x2. The fully connected system consisted of 3 layers with 1024, 512, 128 neurons respectively, with a dropout keep probability of 0.2, 0.2, and 1. Each layer had ELU activation function with He uniform initialization. The output layer had one unit, ReLU activation function with He uniform initialization and batch normalization. Batch size of 128 was chosen. ResNet50_f_age consists of a fully-connected system of 5 layers with 512, 512, 512, 256, 128 units with dropout with keep probability of 0.5, 0.3, 0.3, 0.3 and 0.5 respectively. Each of the layers contains batch normalization and has Scaled Exponential Linear Unit (SELU) as the activation function. Like previously, for SENet50_f_age we kept the same model as for ResNet50_f_age.

**Training & Testing.** In order to save training time each set was separately forward passed via each model to get corresponding 9 Numpy ndarrays as extracted input feature vectors and saved. Since the faces were already aligned and cropped no further preprocessing was carried out and input dimensions are kept same as original RGB photos i.e., 200 x 200 x 3. For gender classification, the loss is binary cross-entropy function. Class weights were also taken into account while training to make up for slight class imbalance as there are roughly 48% female and 52% male in the both the training and validation set. For age estimation, being a regression task, the loss function was mean squared error. The optimizer used in both cases is the AMSGrad variant of Adam [15] with an initial learning rate of 0.001 which is halved in the ending phase of training for better convergence. The choice of optimizer was based on the experiments carried out while training our custom CNN architecture and theory [15].

## 4 Evaluation

The performance of the age estimation algorithms is evaluated based on the closeness of the predicted value to the actual value. The metrics widely used for the age estimation as a regression task is the mean absolute error or MAE which captures the average magnitude of error in a set of predictions. MAE calculates the absolute error between actual age and predicted age as defined by the equation (1).

$$MAE = \frac{1}{n} \sum_{j=1}^{n} |y_j - \hat{y}_j| \qquad (1)$$

Where n is the number of testing samples, $y_j$ denotes the ground truth age and $\hat{y}_j$ is the predicted age of the j-th sample.

For classification tasks (age and gender), the evaluation metric used was accuracy which denotes the fraction of correctly classified samples over the total number of samples.

## 5 Experimentation and Results

In this section we summarize the results obtained via the extensive experiments performed in the study and compare different methods from work of other researchers.

### 5.1 Deep CNNs

We experiment our models in 3 distinct steps. Each successive step uses the model performing the best in the previous step.

First, we tried two of the most popular layer types for convolutional layers. We trained and tested the performance of all - age estimation, age classification and gender classification on 2 types of fundamental convolutional layers - the simple convolutional layer (Conv2D) and separable convolutional layer (Separable Conv2D) with spatial dropout being present in both cases, for increased regularization effect. Rest all hyper parameters were kept the same.

**Table 5.** Comparison of Layer Type

| Layers | Age Estimation (MAE) | Age Classification (Accuracy) | Gender Classification (Accuracy) |
|---|---|---|---|
| Conv2D | 6.098 | 76.718 | 90.426 |
| Separable Conv2D | **6.080** | **78.279** | **91.269** |

It is apparent that separable convolution coupled with spatial dropout (in the convolutional layers) helped the model in converging faster and generalize better. This is because, separable convolutions consist of first performing a depth wise spatial convolution (which acts on each input channel separately) followed by a pointwise convolution which mixes the resulting output channels. Basically, separable convolutions factorize a kernel into 2 smaller kernels, leading to lesser computations, thereby helping the model to converge faster.

Then we experimented with other arguments associated with the namely the type of weight initialization and weight constraints which determine the final weights of our model and hence its performance. Table 6 summarizes the results of this experiment.

**Table 6.** Comparison of models based on the arguments of the best performing layer

| Layers Configuration | Age Estimation (MAE) | Age Classification (Accuracy) | Gender Classification (Accuracy) |
|---|---|---|---|
| Separable Conv2D + Spatial dropout + Xavier uniform initialization | 6.08 | 78.279 | 91.269 |
| Separable Conv2D + Spatial dropout + He uniform initialization | **5.91** | **79.122** | 89.287 |
| Separable Conv2D + Spatial dropout + He uniform initialization + max norm weight constraint | 6.19 | 72.163 | **94.517** |

'He' initialization resulted in better performance when ReLU activation function was used than Xavier initialization.

Again, for each task we chose the configuration that gave the best result on the validation set and tried a bunch of different optimizers in order to maximize performance. Optimizer plays a very crucial role in deciding the model performance as it decides the converging ability of a model. Ideally, we want to have an optimizer that not only converges to the minimum fast, but also helps the model generalize well.

**Table 7.** Effect of various optimizers on results

| Optimizer | Age Estimation (MAE) | Age Classification (Accuracy) | Gender Classification (Accuracy) |
|---|---|---|---|
| Adam | 5.916 | **79.122** | **94.517** |
| Adamax | **5.673** | 78.279 | 91.269 |
| SGD | 6.976 | 70.012 | 89.877 |
| SGD + Momentum (0.9) | 8.577 | 77.098 | 89.624 |
| Nadam | 5.858 | 78.405 | 89.709 |

These results show Adam and its variant (Adamax) provide the best results. Adam and its variants were observed to converge faster. On the other hand, it was observed that models trained using SGD were learning very slowly and saturated much earlier especially when dealing with age.

### 5.2 Transfer Learning

Table 8 compares the performance based on the different extracted features, on which our models were trained.

**Table 8.** Comparison based on feature extractors

| Feature Extractor | Age Estimation (MAE) | Gender Classification (Accuracy) |
|---|---|---|
| VGG_f | 4.86 | 93.42 |
| ResNet50_f | 4.65 | 94.64 |
| **SENet50_f** | **4.58** | **94.94** |

It is clear that the features extracted using SENet-50_f performed best for both the tasks compared to ResNet50_f and VGG_f even though we trained more layers for VGG_f.

In the study [7], a linear regression model and ResNeXt-50 (32×4d) architecture was trained from scratch on the same dataset for age estimation using Adam. In another study [6], various state-of-the-art models pre-trained on ImageNet were used where the authors trained two new layers while freezing the deep CNN layers which acted as feature extractor followed by fine-tuning the whole network with a smaller learning rate later using SGD with momentum. Both studies had their models evaluated on 10% size of the dataset, utilizing remaining for training or validation.

**Table 9.** Comparison with others' work

| Method | Age Estimation (MAE) |
|---|---|
| Linear Regression [7] | 11.73 |
| ResNet50 [6]* | 9.66 |
| Inceptionv3 [6]* | 9.50 |
| DenseNet [6]* | 9.19 |
| ResNeXt-50 (32x4d) [7] | 7.21 |
| Best Custom CNN (ours) | 5.67 |
| VGG_f_age (ours) | 4.86 |
| ResNet50_f_age (ours) | 4.65 |
| **SENet50_f_age (ours)** | **4.58** |

* Cropped according to the detected face image using Haar Cascade Face detector [16].

Since we got best performance, from features extracted via SENet50_f, for both tasks of age and gender classification in Table 10 and Table 11, we further provide baseline performance on them for various machine learning algorithms on the same splits of the dataset. Validation set is not used since we haven't tuned these models, default hyper parameters of Sci-kit learn and XGBoost libraries have been used for this.

**Table 10.** Untuned baseline for Gender Classification

| Method | Train (Accuracy) | Test (Accuracy) |
|---|---|---|
| Decision Tree | 99.86 | 59.26 |
| Linear SVC | 96.13 | 91.44 |
| Logistic Regression | 97.38 | 92.11 |
| Gradient Boosted Trees | 95.15 | 93.38 |
| XGBoost | 95.03 | 93.80 |
| Linear Discriminant Analysis | 95.30 | 94.39 |
| **SVC (kernel = rbf)** | **97.13** | **94.64** |

**Table 11.** Untuned Baseline for Age estimation

| Method | Train (MAE) | Test (MAE) |
|---|---|---|
| Decision Tree | 0.05 | 9.86 |
| Gradient Boosted Trees | 4.97 | 6.17 |
| XGBoost | 5.00 | 5.89 |
| Random Forest | 1.91 | 5.75 |
| Linear Regression | 4.93 | 5.61 |
| Linear SVR | 4.85 | 5.58 |
| **SVR (kernel = rbf)** | **4.85** | **5.49** |

Clearly, even simple linear regression outperformed training our custom CNN model for age estimation and logistic regression came remarkably close to the custom CNN architecture for gender classification on the features extracted using SENet50_f.

As expected, our model performs relatively poorly while predicting ages for people above 70 years of age. This is quite evident from Table 2. where it can be seen that there are only 5.78 % images in the dataset belonging to people above 70 (albeit the dataset is quite evenly balanced when it comes to gender). We believe much better results can be attained using a more balanced and larger dataset.

## 6    Conclusion

Inspired by the recent developments in this field, in this paper we proposed two ways to deal with the problem of age estimation, age and gender classification - a custom CNN architecture and transfer learning based pre-trained models. These pre-trained models helped us combat overfitting to a large extent. It was found that our models generalized very well with minimal overfitting, when tested on real-life images.

We plan to extend our work on a larger and more balanced dataset with which we can study biases and experiment with more things in order to improve the generalizability of our models. In future research, we hope to use this work of ours as a platform to improvise and innovate further and contribute to the deep learning community.